\documentclass[conference]{IEEEtran}
\usepackage{times}
\usepackage{epsfig}
\usepackage{graphicx}
\usepackage{amsmath}
\usepackage{amssymb}
\usepackage{algorithmic}
\usepackage{subfigure}
\usepackage{caption}
\usepackage{float}
\usepackage[ruled,vlined]{algorithm2e}
\usepackage{multirow,tabularx}
\usepackage{amsmath}
\usepackage{amssymb}

\usepackage{multirow}
\usepackage{rotating}
\usepackage{color}

\usepackage{xspace}

\DeclareRobustCommand{\etal}{\textit{et al.}\@\xspace}

\makeatletter
\DeclareRobustCommand{\etc}{%
    \@ifnextchar{.}%
        {\textit{etc}}%
        {\textit{etc.\@\xspace}}%
}

\ifCLASSINFOpdf
\else
\fi
\begin{document}
\title{Magnifying Subtle Facial Motions for Effective 4D Expression Recognition}

\author{\IEEEauthorblockN{Qingkai Zhen}
\IEEEauthorblockA{School of Computer Science and \\Engineering, Beihang University, China\\
Email: qingkai.zhen@buaa.edu.cn}
\and
\IEEEauthorblockN{Di Huang}
\IEEEauthorblockA{School of Computer Science and \\Engineering, Beihang University, China\\
Email: dhuang@buaa.edu.cn}
\and
\IEEEauthorblockN{Yunhong Wang}
\IEEEauthorblockA{School of Computer Science and \\Engineering, Beihang University, China\\
Email: yhwang@buaa.edu.cn
}
\and
\IEEEauthorblockN{Hassen Drira}
\IEEEauthorblockA{Institut Mines-T\'{e}l\'{e}com/T\'{e}l\'{e}com Lille,\\ CRIStAL (UMR CNRS 9189), France\\
Email: hassen.drira@telecom-lille.fr}
\and
\IEEEauthorblockN{Boulbaba Ben Amor}
\IEEEauthorblockA{Institut Mines-T\'{e}l\'{e}com/T\'{e}l\'{e}com Lille,\\ CRIStAL (UMR CNRS 9189), France\\
Email: boulbaba.benamor@telecom-lille.fr}
\and
\IEEEauthorblockN{Mohamed Daoudi}
\IEEEauthorblockA{Institut Mines-T\'{e}l\'{e}com/T\'{e}l\'{e}com Lille,\\ CRIStAL (UMR CNRS 9189), France\\
Email: mohamed.daoudi@telecom-lille.fr}
}


%Telephone: (800) 555--1212, Fax: (888) 555--1212}


\maketitle

\begin{abstract}
In this paper, an effective pipeline to automatic 4D Facial Expression Recognition (\textit{4D FER}) is proposed. It combines two growing but disparate ideas in Computer Vision -- computing the spatial facial deformations using tools from Riemannian geometry and magnifying them using temporal filtering. The flow of 3D faces is first analyzed to capture the spatial deformations based on the recently-developed Riemannian approach proposed in \cite{benamor2014}, where registration and comparison of neighboring 3D faces are led jointly. Then, the obtained temporal evolution of these deformations are fed into a magnification method in order to amplify the facial activities over the time. The latter, main contribution of this paper, allows revealing subtle (hidden) deformations which enhance the emotion classification performance. We evaluated our approach on BU-4DFE dataset, the state-of-art 94.18\% average performance and an improvement that exceeds 10\% in classification accuracy, after magnifying extracted geometric features (deformations), are achieved.
\end{abstract}


% \fi
\IEEEpeerreviewmaketitle

\section{Introduction}\label{Intro}

Facial expressions are important non-verbal ways for human beings to communicate their feeling and affective states. In recent years, as a major topic of affective computing, Facial Expression Recognition (\textit{FER}) has attracted increasing interests due to its potential in many applications, such as psychological analysis \cite{Meng2013}, transport security (driver fatigue), computer graphics, human-machine interaction, animation of 3D avatars \textit{etc}.

%Ekman and Friesen \cite{Ekman} have defined six universal facial expressions which are \textit{Anger (AN)}, \textit{Disgust (DI)}, \textit{Fear (FE)}, \textit{Happiness (HA)}, \textit{Sadness (SA)}, \textit{Surprise (SU)}, consistent across different races and cultures. Inspired by these fundamental discoveries, research in \textit{FER} was extensively developed targeting this classification.

Sun and Yin, the pioneers of 4D \textit{FER}, tracked the change of a generic deformable model to
extract a \textit{Spatio-Temporal} (\textit{ST}) descriptor of the face from dynamic sequences of 3D
scans \cite{SunECCV2008}. The vertex flow tracking was applied to each frame
to form a set of motion trajectories of the 3D face video.
The spatio-temporal features and two-dimensional \textit{HMM} were used for classification.
In \cite{Sun2010}, they proposed a tracking-model-based approach for vertex registration and
motion trajectory estimation in 4D \textit{FER}. The 2D intermediary generated through conformal mapping
and a generic model adaptation algorithm were employed to establish the correspondence across frames.
\textit{ST-HMM} which incorporates 3D surface characterization was utilized to learn the spatial and
temporal information of faces.
Canavan \textit{et al} \cite{Canavan2012} described the 3D dynamic surface by the surface curvature-based
shape-index information, and the surface features are characterized in local regions along the temporal axis.
A dynamic curvature descriptors is constructed from local regions as well as temporal domains, and 3D tracking model
based method was applied to locate the local regions across 3D dynamic sequences.
Sandbach \textit{et al}  \cite{Sandbach2011FG} exploited 3D motion-based features (\textit{Free-Form Deformation, FFD})
between neighboring 3D facial geometry frames for \textit{FER}. A feature selection step was applied to localize the features
of each of the onset and offset segments of the expression. The \textit{HMM} classifier was used to model the full temporal
dynamics of each expression. In their another work \cite{Sandbach2012}, the entire expressive sequence is modelled to contain an \textit{Onset} followed by an \textit{Apex} and an \textit{Offset}. Feature selection methods are applied in order to extract features for each of the onset and offset segments of the expression. These features are then used to train
\textit{GentleBoost} classifiers and build an \textit{HMM} in order to model the full temporal dynamics of the expression.

Ben Amor \textit{et al} \cite{benamor2014, DriraADSB12, DaoudiDAB13} presented the facial by collections of radial curves,
then Riemannian shape analysis was applied to quantify dense deformations and extract motion from successive 3D frames.
Two different classification schema were performed, a \textit{HMM-based} classifier and a mean deformation-based classifier.
Xue \textit{et al} \cite{Xue2015} extracted local depth patch-sequences from consecutive expression frames based on the automatically detected facial landmarks. Three dimension discrete cosine transform (\textit{3D-DCT}) is then applied on these patch-sequence to extract spatio-temporal features for facial expression dynamic representation.
Berretti \textit{et al} \cite{Berretti2013} presented a fully-automatic and real-time approach for 4D \textit{FER}.
A set of 3D facial landmarks were automatically detected firstly, the local characteristics of the face around those landmarks
and their mutual distances were used to model the facial deformation.
The work of Fang \textit{et al} \cite{iccvwFangZSK11,journalsivcFangZOSK12} emphasized 4D face data registration and dense corresponding between 3D meshes along the temporal line, and a variant of \textit{Local Binary Patters on Three Orthogonal Plane} (\textit{LBP-TOP}) was introduced for static and dynamic feature extraction to predict expression labels.

Even though the performance of \textit{FER} has been substantially boosted by 4D data in recent years, there still exist an unsolved problem, that is the reputed similar expressions are difficult to distinguish since the facial deformations are sometimes really slight \cite{conffgrYang0WC15}. To handle this issue, this paper present a novel and effective approach to
amplify the subtle facial deformation, the contribution of the paper are two-folds:
\begin{itemize}
  \item A comprehensive pipeline of spatio-temporal processing for effective facial expression recognition from 4D data.
  \item A method to amplify subtle movements on facial surfaces which contributes to distinguish similar expressions.
\end{itemize}

The rest of the paper is structured as follows. In section \ref{GeometricFeature} the background of the used feature is introduced. The magnification of subtle facial deformation is introduced in Section \ref{Magnification}. Our comprehensive experimental study is presented in Section \ref{ExperimentsResult}, followed in Section \ref{Concl} we conclude the paper.

%%%%%%%%%%%%%%%%%%%%%%%%%%%%%%%%%%%%%%%%%%%%%%%% Section 2. Extract Geodesic Features.
\section{Background on Dense Scalar Fields} 
\label{GeometricFeature}
%%%%%%%%%%%%%%%%%%%%%%%%%%%%%%%%%%%%%%%%%%%%%%%% 
\begin{figure}[!ht]
\centering
\includegraphics[width = \linewidth]{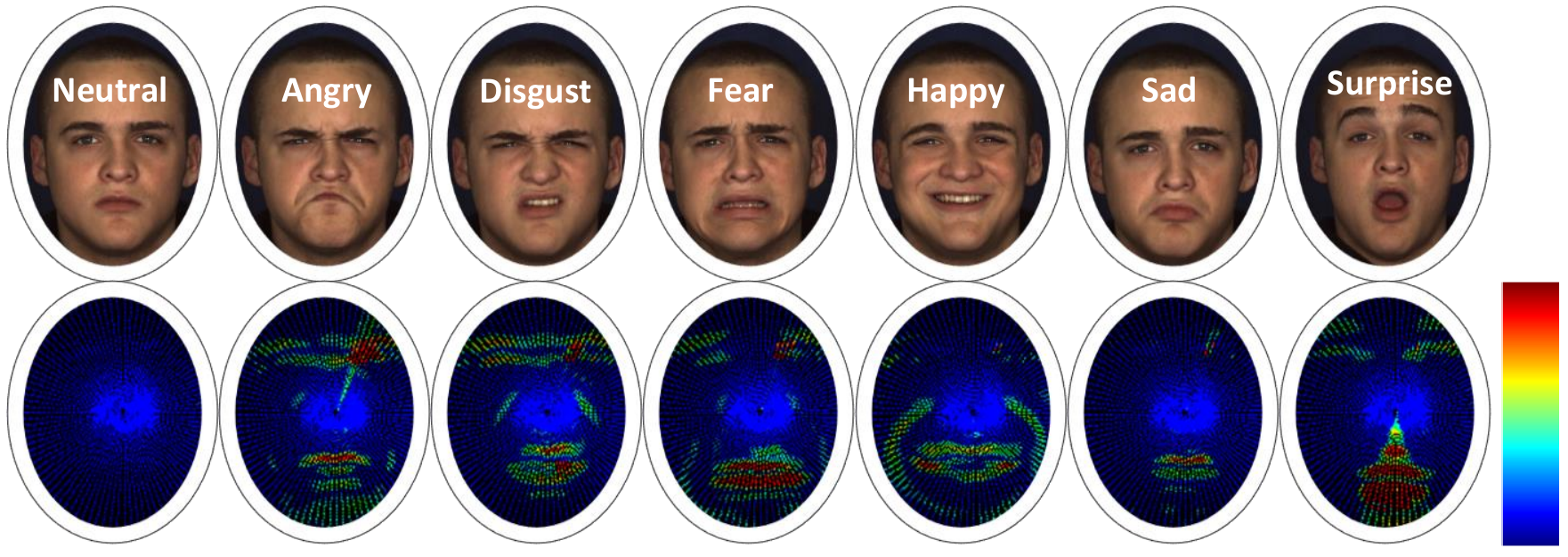}
\caption{Top row: facial texture images of an individual with different expressions. Bottom row: facial deformations in Riemannian space. Where, warm colors are associated to the high $\chi$ and correspond to facial regions with high deformations, cold colors reflect the most static parts of the 3D face. }
\label{DSF_Compute}
\end{figure}
%%%%%%%%%%%%%%%%%%%%%%%%%%%%%%%%%%%%%%%%%%%%%%%% 

Following the geometric approach recently-developed in \cite{benamor2014}, we represent 3D facial surfaces by collections of radial curves emanating from the tip of the nose. It is a new parameterization imposed for 3D face analysis, registration, comparison, \etc The amount of deformation from one shape into another (across the 3D video) is computed using tools from differential geometry through analyzing shapes of 3D radial curves as explained in the following. In the pre-processing step, the 3D mesh in each frame is first aligned to the first one and then cropped. The facial surfaces are then approximated by indexed collections of radial curves $\beta_{\alpha}$, where the index $\alpha$ denotes the angle formed by the curve with respect to a reference radial curve. These curves are then uniformally resampled. Given a radial curve $\beta$ of the face with an arbitrary orientation $\alpha\in [0,2\pi]$, it can be parameterized as $\beta: I \to \mathbb{R}^3$, with $I = [0,1]$, and mathematically represented using the SRVF, denoted by $q(t)$, according to:
%$q(t) = \frac{\dot{\beta}(t)}{\sqrt{\| \dot{\beta}(t) \|}}, t \in I.$

\begin{equation}\label{SRVF}
q(t) = \frac{\dot{\beta}(t)}{\sqrt{\| \dot{\beta}(t) \|}}, t \in I.
\end{equation}

%%%%%%%%%%%%%%%%%%%%%%%%%%%%%%%%%%%%%%%%%%%%%%%%%%%%%%%%%%%%%%%%%%%%%%
\begin{figure*}[!ht]
\centering
\subfigure[]{
\includegraphics[width = 0.7\linewidth]{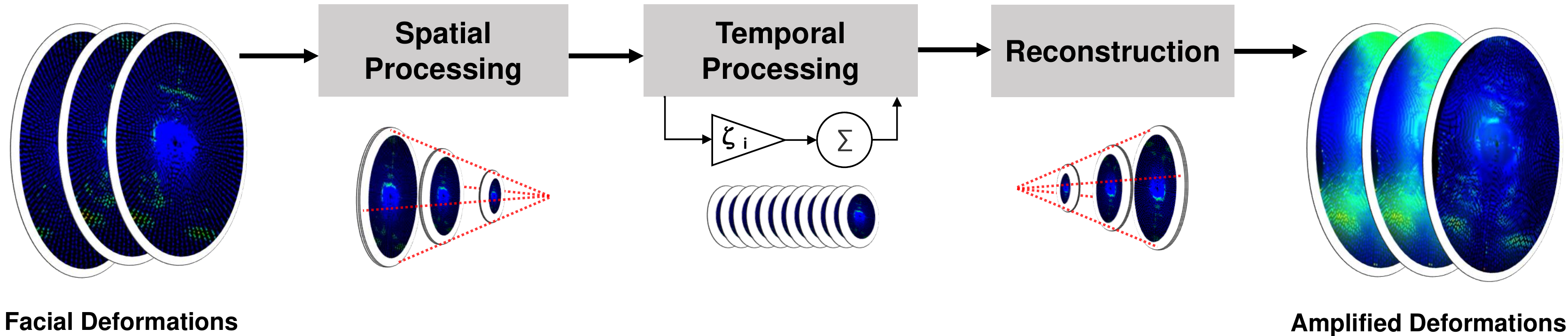}
\label{GaussianPyramid}
}
\subfigure[]{
\includegraphics[width = 0.25\linewidth]{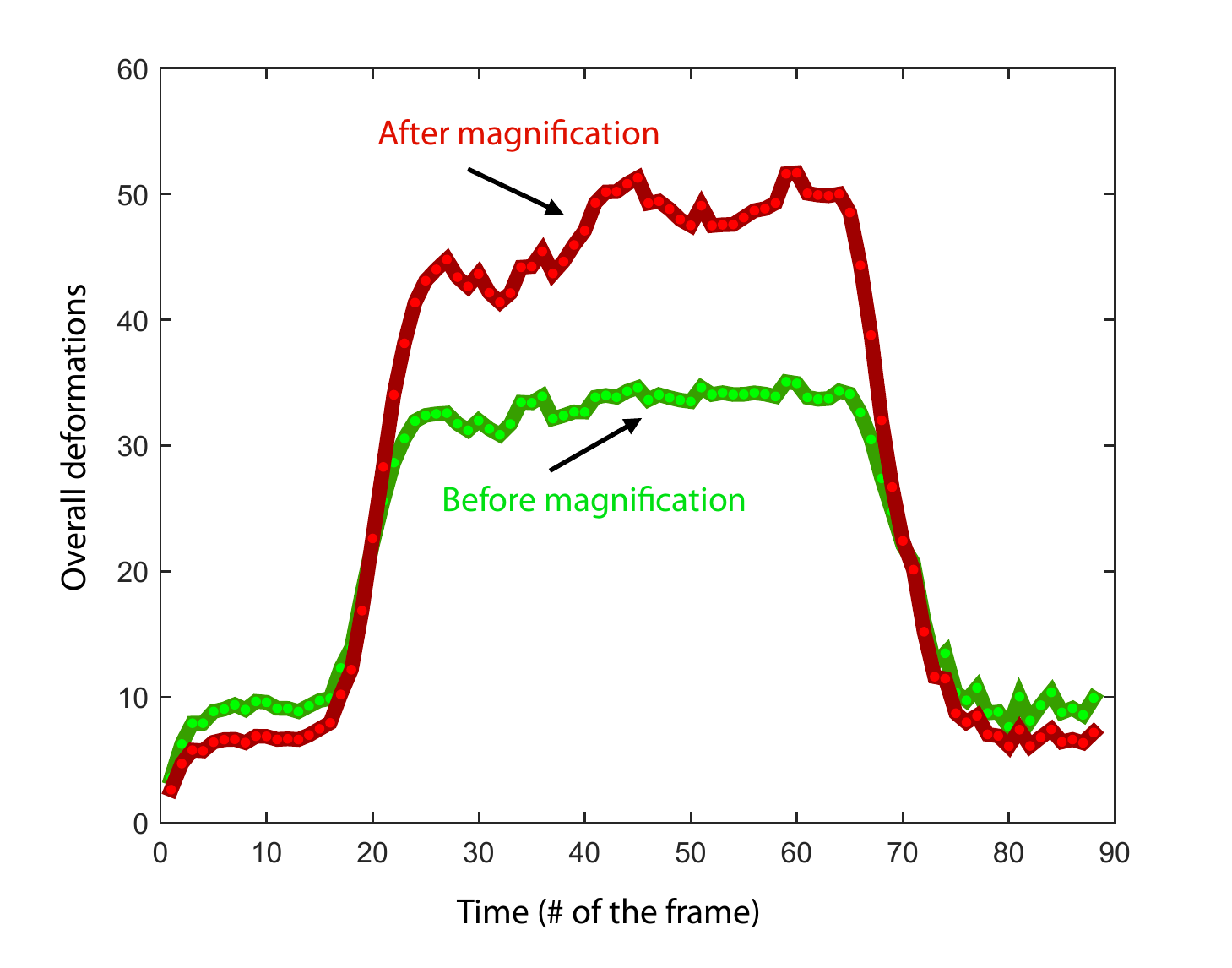}
\label{DSF_Amplified_DSF}
}
\label{GaussianPyramid_Amplified}
\caption{(a) Overview of 3D video magnification. The original facial deformation features are first decomposed into different spatial frequencies, and the temporal filter is applied to all the frequency bands. The filtered spatial bands are then amplified by a given factor $\zeta$, added back to the original signal, and collapsed to the output sequence. (b) An example of facial expression deformation (norm of the velocity vector) before (green) and after (red) magnification.}
\end{figure*}
%%%%%%%%%%%%%%%%%%%%%%%%%%%%%%%%%%%%%%%%%%%%%%%%%%%%%%%%%%%%%%%%%%%%%%

This special representation has the advantage of capturing the shape of the curve and makes the calculus simpler. While there are several ways to analyze shapes of curves, an elastic analysis of the parametrized curves is particularly appropriate in our application – face analysis under facial expression variations. This is because (1) such analysis uses the square-root velocity function representation which allows us to compare local facial shapes in presence of deformations, (2) this method uses a square-root representation under which the elastic metric is reduced to the standard $\ensuremath{\mathbb{L}^2}$ metric and thus simplifies the analysis, (3) under this metric the group of re-parametrization acts by isometry on the curves manifold, thus a Riemannian re-parametrization metric can be set between two facial curves. Shown in Fig. \ref{DSF_Compute} are examples of apex frames taken from the 3D videos of the  BU-4DFE dataset as well as the dense 3D deformations computed with respect to the neutral frame. Let us define the space of the SRVFs as 
\begin{equation}\label{LabelC}
\mathcal{C} = \{ q: I \to \mathbb{R}^3, \| q \| =1 \} \subset \ensuremath{\mathbb{L}^2}(I,\mathbb{R}^3), 
\end{equation}

with$ \| \cdot \|$ indicating the $\ensuremath{\mathbb{L}^2}$ norm. With the $\ensuremath{\mathbb{L}^2}$ metric on its tangent space, $\mathcal{C}$ becomes a Riemannian manifold. Basically, with this parametrization each radial curve is represented on the manifold $\mathcal{C}$ by its SRVF. Accordingly, given the SRVF $q_{1}$ and $q_{2}$ of two curves, the shortest path $\psi^{\ast}$ on the manifold $\mathcal{C}$ between $q_{1}$ and $q_{2}$ (called geodesic path) is a critical point of the following energy function: 
%$E(\psi) = \frac{1}{2} \int \| \dot{\psi}(\tau)\|^2 d \tau,$
\begin{equation}\label{energyfunction}
E(\psi) = \frac{1}{2} \int \| \dot{\psi}(\tau)\|^2 d \tau,
\end{equation}
where $\psi$ denotes a path on the manifold $\mathcal{C}$ between $q_{1}$ and $q_{2}$, $\tau$ is the parameter for traveling along the path $\psi$, $\dot \psi(\tau) \in T_{\psi(\tau)}(\mathcal{C})$ is the tangent vector field on the curve $\psi(\tau) \in \mathcal{C}$, and $\|\cdot\|$ denotes the $\ensuremath{\mathbb{L}^2}$ norm on the tangent space. Since elements of $\mathcal{C}$ have a unit $\ensuremath{\mathbb{L}^2}$ norm, $\mathcal{C}$ is an hypersphere in the Hilbert space $\ensuremath{\mathbb{L}^2}(I,\mathbb{R}^3)$. As a consequence, the geodesic path between any two points $q_{1}$ and $q_{2} \in \mathcal{C}$ is given by the minor arc of the great circle connecting them. The tangent vector field on this geodesic between the curves $\beta_1$ and $\beta_2$ making the angle $\alpha$ with the reference curve is parallel along the geodesic and one can represent it with the initial velocity vector (called also shooting vector), %given in  Eqn. \ref{AnotherFormat}, 
without any loss of information. 
%$\frac{d\psi^{\ast}_{\alpha}}{d\tau}_{\arrowvert_{\tau = 0}} = \frac{\theta}{\sin(\theta)}(q_{2}-\cos(\theta)q_{1}),\ (\theta \neq 0).$

\begin{equation}\label{AnotherFormat}
\frac{d\psi^{\ast}_{\alpha}}{d\tau}_{\arrowvert_{\tau = 0}} = \frac{\theta}{\sin(\theta)}(q_{2}-\cos(\theta)q_{1}),\ (\theta \neq 0).
\end{equation}

where $\theta = d_{\mathcal{C}}(q_{1},q_{2}) = cos^{-1}(\langle q_{1},q_{2} \rangle)$ represent the length of the geodesic path connecting $q_{1}$ to $q_{2}$. In practice, the curves are re-sampled to a discrete number of points, say $T$, and the face is approximated by a collection of $\arrowvert \Lambda \arrowvert$ curves. The norm of the quantity 
%given in Eqn. \ref{AnotherFormat} 
at each discrete point $r$ is computed to represent the amount of 3D deformations at each point of the surface parameterized by the pair $(\alpha,r)$, termed Dense Scalar Fields (or \textit{DSFs}). The final feature vector is of size $T\times \arrowvert \Lambda \arrowvert$. We will refer to this quantity at a given time $t$ of the 3D video by $\chi(t)$ (see bottom row of Fig. \ref{DSF_Compute} for illustrations). It provides the amplitude of the deformations between two facial surfaces in a dense way.

%%%%%%%%%%%%%%%%%%%%%%%%%%%%%%%%%%%%%%%%%%%%%%%%%%%%%%%%
\section{Subtle Facial Deformation Magnification}
\label{Magnification}

As described in Section \ref{GeometricFeature}, $\chi$ reveals the shape difference of two facial surfaces by deforming one mesh into another through an accurate registration step. However, there exist another challenge to capture certain facial movements, especially the slight ones, with low spatial amplitude, reflected by the limited performance in distinguishing similar 3D facial expressions in the literature. To solve this problem, we propose a novel approach to highlight the subtle geometry changes of the facial surface in $\chi$  by adapting the Eulerian spatio-temporal processing \cite{Wu12Eulerian} to the 3D domain. This method and its application to 3D face videos are presented in the subsequent.  The Eulerian spatio-temporal processing was introduced for motion magnification in 2D videos, and has proved its effectiveness in \cite{Wu12Eulerian}.  Its basic idea is to amplify the variation of pixel values over time, in a spatially-multiscale manner, without explicitly estimating motion but rather exaggerating motion by amplifying temporal color changes at fixed positions. It relies on a linear approximation related to the brightness constancy assumption that forms the basis of the optical flow algorithm. However, the case is not that straightforward in 3D, because the vertex correspondence across frames cannot be achieved as easy as that in 2D. Fortunately, during the computation of $\chi$, such correspondence is established by surface registration and remeshing. We can thus adapt Eulerian spatio-temporal processing to 3D face video. We take into account the values of the time series $\chi$ at any spatial location and highlight the differences in a given temporal frequency band of interest. It thus combines spatial and temporal processing to emphasize subtle changes in a 3D face video.

The process is illustrated in Fig. \ref{GaussianPyramid}. Specifically, the video sequences are first decomposed into different spatial frequency bands by Gaussian pyramid, and these bands might be magnified differently.
We consider that the time series correspond to the values of $\chi$ on the mesh surfaces in a frequency band and apply a band pass filter to extract the frequency bands of interest. The temporal processing, $\mathfrak{T}$, is uniform for all spatial levels, and for all $\chi$ within each level. We then multiply the extracted band passed signal by a magnification factor $\zeta$, and add the magnified signal to the original and collapse the spatial pyramid to obtain the final output.

%%%%%%%%%%%%%%%%%%%%%%%%%%%%%%%%%%%%%%%%%%%%%%%% Algorithm
\begin{algorithm}
\label{MagnificationAlgorithm}
\caption{Online 3D Deformation Magnification}
\KwIn{$\chi$, $l$-Gaussian pyramid levels, $\zeta$-amplification factor, $\mathcal{\xi}$-sample rate, $\gamma$-attenuation rate, $\mathfrak{f}$-video frequency}
\textbf {Step1. Spatial Processing}

\For {$i = 1; i \le n $}
  {
   $\mathfrak{D}$($i, :, :, :$) = decompose the $\chi(i)$, with $l$ level Gaussian pyramid.
  }

  \textbf {Step2. Temporal Processing}

  $\mathfrak{S}$ = $\mathfrak{T}$($\mathfrak{D}$, $\mathfrak{f}$, $\mathcal{\xi}$)

  \textbf {Step3. Magnification}

  \For {$i = 1; i \le 3$}
  {
   $\mathfrak{S(:,:,:,i)}$ = $\mathfrak{S(:,:,:,i)} \ast \zeta \ast \gamma$
  }

   \textbf {Step3. Reconstruction}

  \For {$i = 1; i \le n$}
  {
   $\hat{\chi}(i)$ = $\mathfrak{S}(i, :, :, :)$ + $\chi(i)$
  }
\KwOut{$\hat{\chi}(t)$}
\end{algorithm}

% %%%%%%%%%%%%%%%%%%%%%%%%%%%%%%%%%%%%%%%%%%%%%%%%%%%%%%%%

For the translational motion of the facial mesh, we express the observed $\chi(s, t)$  value with respect to a displacement function $\delta(t)$, such that $\chi(s,t) = \chi(s) + \delta(t)$ and $\chi(s,0) = \chi(s)$. By using a first-order Taylor series expansion, at time $t$, $\chi(s+\delta(t))$ in a first-order Taylor expansion about $s$, as
\begin{equation}\label{A}
\chi(s,t) \thickapprox \chi(s) + \delta(t) \frac{\partial{\chi(s)}}{\partial {s}}
\end{equation}
Let $\phi(s,t)$ be the result of applying a broadband temporal band pass filter to $\chi(s)$ at each position $(s)$.
Assume that the motion signal $\delta(t)$ is within the pass band of the temporal band pass filter.
\vspace{-2mm}
\begin{equation}\label{B}
\phi(s,t) = \delta(t)\frac{\partial{\chi(s)}}{\partial{s}}
\end{equation}
Amplify the band pass signal by factor $\zeta$ and add it back to $\chi(s)$.
\begin{equation}\label{C}
\hat{\chi}(s,t) = \chi(s,t) + \zeta \phi(s,t)
\end{equation}
By combining Eqn. \ref{A}, \ref{B}, and \ref{C}, we reach
\begin{equation}\label{D}
\hat{\chi}(s,t) \thickapprox \chi(s) + (1+\zeta)\delta(t)\frac{\partial{\chi(s)}}{\partial{s}}
\end{equation}
Assuming that the first-order Taylor expansion holds for the amplified larger perturbation $(1+\zeta)\delta(t)$, the motion magnification of 3D face video can be simplified as follows:
\begin{equation}\label{D}
\hat{\chi}(s,t) \thickapprox \chi(s + (1+\zeta)\delta(t))
\end{equation}
This shows that the spatial displacement $\delta(t)$ of the $\chi(s)$ at time $t$, is amplified to a magnitude of $(1+\zeta)$.
Sometimes $\delta(t)$ is not entirely within the pass band of the temporal filter.
In this case, let $\delta_{k}(t)$, indexed by $k$, represent the different temporal spectral components of $\delta(t)$.
This results in a band pass signal,
\begin{equation}
\phi(s,t) = \sum_{k} \gamma_{k}\delta_{k}(t)\frac{\partial{\chi(s)}}{\partial{s}}
\end{equation}
where $\gamma$ is an attenuation factor.
Temporal frequency dependent attenuation can be equivalently interpreted as a frequency-dependent motion magnification factor, $\zeta_{k} = \gamma \zeta$, and the amplified output signal
\begin{equation}
\hat{\chi}(s,t) \thickapprox \chi(s + \sum_{k}(1+\zeta{k})\delta_{k}(t))
\end{equation}
this procedure,  and Fig. \ref{DSF_Amplified_DSF} displays an example of facial deformation trajectory before (green) and after (red) magnification.

%%%%%%%%%%%%%%%%%%%%%%%%%%%%%%%%%%%%%%%%%%%%%%%% Section 5. Experimental Results.
\section{Experimental Results}
\label{ExperimentsResult}

\subsection{Dataset Description and Experimental Settings}

The BU-4DFE dataset \cite{SunECCV2008} is a dynamic 3D facial expression dataset which consists of 3D facial sequences of $58$ females and $43$ males. It includes in total $606$ 3D sequences according to the $6$ universal expressions.
Each 3D sequence captures a facial expression at a rate of $25$ fps (frames per second) and lasts approximately $3$-$4$ seconds.

In our experiments, at a time $t$, the 3D face model $f^t$ is approximated by a set of $200$ elastic radial curves originating from the tip of the nose, a total of 50 sampled vertices on each curve is considered. Based on this parameterization, the 3D face shapes along the video sequence are compared to a reference frame $f^0$  to derive the $\chi(t)$ at each time $t$. Then, within the spatial processing step, a Gaussian pyramid decomposition is used to decompose $\chi$ into $4$ band levels. Finally, a temporal processing to all the bands is applied. The factor $\zeta$ is set to $10$, the sample rate $\mathcal{\xi}$ is set to $25$, $\mathfrak{f} \in [0.3, 0.4]$, and the attenuation rate $\gamma$ is set to $1$. Our experiments are conducted on the following sub-pipelines -- (1) the whole video sequence (denoted by \textit{WV}), (2) the magnified whole video sequence (denoted as \textit{MWV}).

A multi-class Support Vector Machine (\textit{SVM}) is considered here where $\bar{\chi}$ is treated as a feature vector to predict the video label in the Subtle Deformation Magnification experiment (using \textit{SVM Classifier}).
We also adopt $\textit{HMM}$ to encode the temporal behavior of each expression, and get the expression type.
To allow fair comparison with previous studies, we randomly select $60$ subjects from the BU-4DFE dataset to perform our experiments under a $10$-fold cross-validation protocol.

\subsection{Classification Performance}
%%%%%%%%%%%%%%%%%%%%%%%%%%%%%%%%%%%%%%%%%%%%%%%%%%%%%%%%%%%%%
\begin{table*}[!ht]
\caption{Average accuracy (and standard deviations) achieved by \textit{SVM} and \textit{HMM} on \textit{whole video sequence} before and after magnification.}
\begin{center}\small
\begin{tabular}{c|c|c}
\hline
\hline
\textbf{Algorithm} & \textbf{Magnification?} & \textbf{Whole Sequence} (\%) \\
\hline
\multirow{2}[6]*{\textit{SVM on $\bar{\chi}$}} & N & $82.49 \pm 3.11$ \\
                    &Y & $93.39 \pm 3.54$  \\
\hline
\multirow{2}[6]*{\textit{HMM on $\chi(t)$}} & N & $83.19 \pm 2.84$ \\
                    &Y & $94.18 \pm 2.46$\\
\hline
\hline
\end{tabular}
\end{center}
\label{Table1}
\end{table*}

\begin{figure*}
\centering
\includegraphics[width=0.8\linewidth]{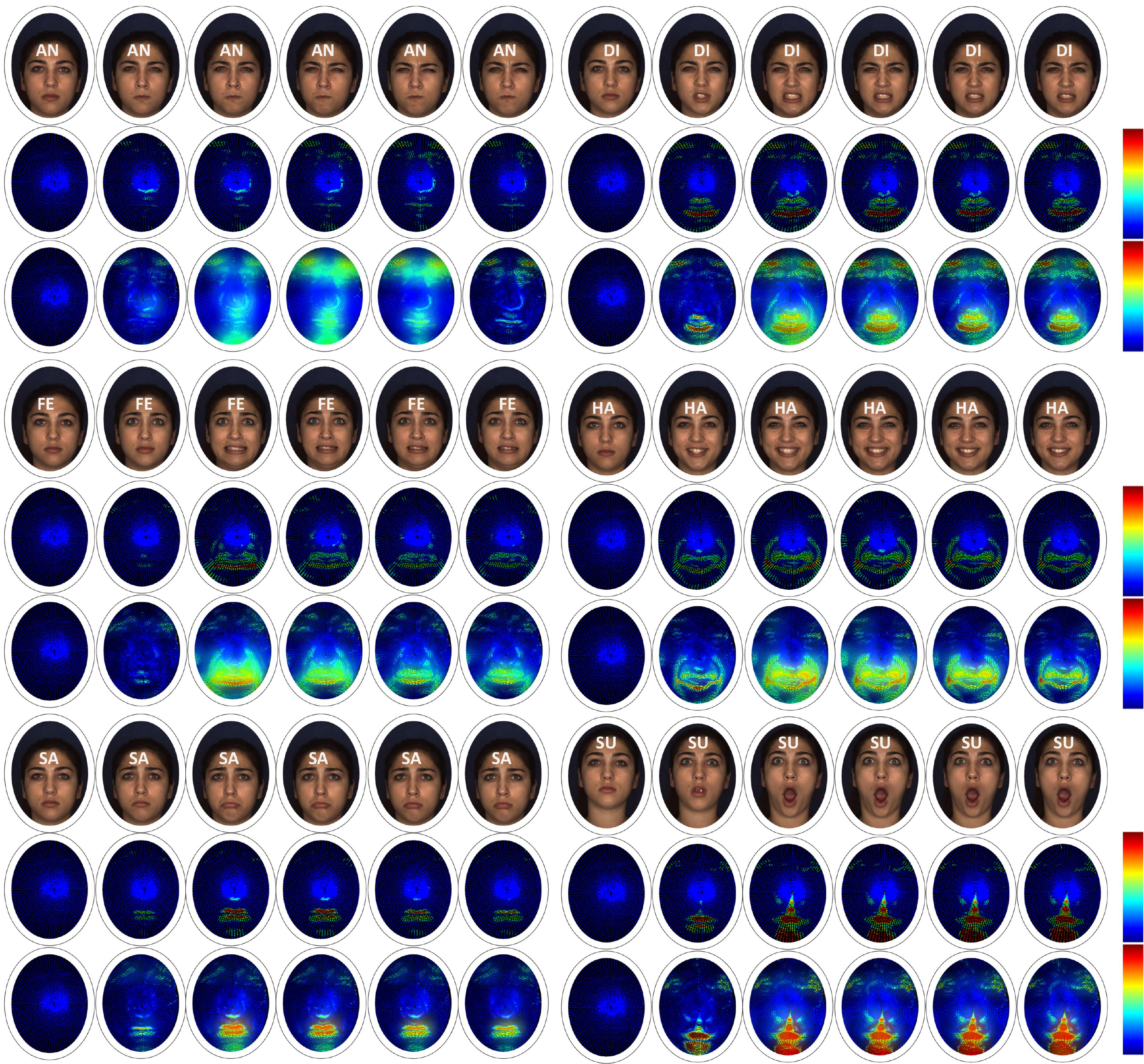}
\caption{Illustrations of the deformation magnification on sequences of the same subject performing six expressions. One can appreciate the magnification effects on 3D deformations compared to those of the original features. From up to bottom, each
row presents the texture image, the original features, and the amplified features, respectively.}
\label{ExpressionsALL}
\end{figure*}

%%%%%%%%%%%%%%%%%%%%%%%%%%%%%%%%%%%%%%%%%%%%%%%%%%%%%%%%%%%%
\begin{table*}[!ht]
\caption{The confusion matrices (\textit{WV}, \textit{MWV}) achieved by using \textit{SVM} and \textit{HMM} classifiers.}
\begin{center}\small
\begin{tabular}{c|c|c|c|c|c|c|c|c|c|c|c|c}
\hline
\hline
\textbf{SVM on $\bar{\chi}$} & \multicolumn{6}{c|}{\textbf{Whole Video (\textit{WV})}}  & \multicolumn{6}{c}{\textbf{Magnified Whole Video (\textit{MWV})}}  \\
\hline
\% & AN & DI & FE & HA & SA & SU & AN & DI & FE & HA & SA & SU \\
\hline
 AN & \textbf{73.86} & 9.18 & 6.49 & 1.75 & 6.11 & 2.51 &\textbf{91.07}&2.73&2.01&1.59&2.08&0.51 \\
\hline
 DI & 8.76 & \textbf{71.27} & 9.29 & 3.51 & 4.84 & 2.21 &2.05&\textbf{92.62}&2.63&1.07&1.38&0.24 \\
\hline
 FE &5.79&5.37&\textbf{73.14}&4.59&5.39&5.61&1.66&1.53&\textbf{92.33}&1.31&1.54&1.62\\
\hline
 HA &0.81 & 1.18 & 2.42 & \textbf{93.6} & 1.08 & 0.88 &0.91&0.88&2.36&\textbf{94.29}&0.97&0.58\\
\hline
SA &2.54&2.27&2.99&1.63&\textbf{88.75}&1.77&1.36&1.22&1.62&0.9&\textbf{93.93}&0.96\\
\hline
SU &0.74&0.88&1.91&0.75&1.38&\textbf{94.32}&0.51&0.61&1.29&0.52&0.96&\textbf{96.11}\\
\hline
\textbf{Average} &\multicolumn{6}{c}{\textbf{82.49 $\pm$ 3.10}}  & \multicolumn{6}{c}{\textbf{93.39 $\pm$ 3.54}}  \\
\hline
%\vspace{0.3cm}
\hline
\textbf{HMM on $\chi(t)$} & \multicolumn{6}{c}{}  & \multicolumn{6}{c}{}  \\
\hline
\% & AN & DI & FE & HA & SA & SU & AN & DI & FE & HA & SA & SU \\
\hline
 AN &\textbf{75.29}&5.88&7.31&1.14&8.17&2.21&\textbf{91.87}&1.91&2.41&0.38&2.69&0.73\\
\hline
 DI &10.42&\textbf{71.55}&11.43&1.82&4.27&0.5&2.11&\textbf{94.22}&2.32&0.29&0.86&0.19\\
\hline
 FE &5.07&6.86&\textbf{73.69}&3.33&8.06&2.99&1.37&1.86&\textbf{92.85}&0.91&2.19&0.81\\
\hline
 HA &0.48&0.87&1.54&\textbf{94.93}&1.81&0.37&0.47&0.77&1.43&\textbf{95.3}&1.67&0.35\\
\hline
 SA &3.71&1.01&4.17&0.65&\textbf{89.19}&1.26&1.84&0.51&2.07&0.33&\textbf{94.61}&0.63\\
\hline
 SU &0.49&0.33&2.79&0.32&1.59&\textbf{94.47}&0.33&0.22&1.89&0.22&1.08&\textbf{96.25}\\
\hline
\textbf{Average} &\multicolumn{6}{c|}{\textbf{83.19 $\pm$ 2.83}}  & \multicolumn{6}{c}{\textbf{94.18 $\pm$ 2.46}}  \\
\hline
\hline
\end{tabular}
\end{center}
\label{FinalResultsSDM1}
\end{table*}
%%%%%%%%%%%%%%%%%%%%%%%%%%%%%%%%%%%%%%%%%%%%%%%%%%%%%%%%%%%%
Table \ref{Table1} provides a first summary of our results, it shows that the magnification procedure an improvement that exceeds 10\% in classification accuracy is achieved.
Before magnification, our approach achieves 82.49\% and 83.19\%  as correct classification rate using \textit{SVM} and \textit{HMM} classifiers respectively, when consider full video. In particular, the \textit{SU} and \textit{HA} sequences are well classified with a high classification rate. This is mainly due to the high intensities and clear patterns of their deformations. However, remaining expressions (\textit{DI}, \textit{FE}, \textit{AN} and \textit{SA}) are harder to distinguish. We believe that two major reasons induce this difficulty: (1) the intra-class variability which makes confusing similar classes such as \textit{DI}/\textit{AN}/\textit{FE}; (2) the low magnitude of the deformations exhibited when performing these expressions. After the magnification, the overall improvement exceeds 10\% both all settings, which highlights the interest of the magnification procedure to reveal invisible deformations. It can be clearly seen from these confusion matrices, a significant improvement in distinguishing the \textit{AN}, the \textit{DI} and the \textit{FE} expressions which have been difficult to recognize before. Table \ref{FinalResultsSDM1} shows the confusion matrices (WV, MWV) achieved by using SVM and HMM classifiers. Fig. \ref{ExpressionsALL} illustrations of the deformation magnification on sequences of the same subject.

\subsection{Comparison with state-of-the-art}

Several research groups have reported \textit{FER} results on the BU-4DFE dataset, however they differ in their experimental settings. In this section, we compare our results with existing approaches when considering these differences.

%%%%%%%%%%%%%%%%%%%%%%%%%%%%%%%%%%%%%%%%%%%%%%%%%%%%%%%%%%%%%%%%%%%%%%%
\begin{table}[!ht]
\caption{A comparative study of the proposed approach with the state-of-the-art on BU-4DFE.}
\begin{center}
\begin{tabularx}{0.5\textwidth}{l|l|l}
\textbf{Method} & \textbf{Experimental Settings} & \textbf{Accuracy} \\
\hline
Sun \etal \cite{SunECCV2008} & 6E, 60S, 10-CV, Win=6 & 90.44\%\\
\hline
Sun \etal \cite{Sun2010} & 6E, 60S, 10-CV, Win=6 & 94.37\% \\%83.7\% \\
\hline
Reale \etal \cite{fgrRealeZY13} & 6E, 100S, --, Win=15 & 76.9\% \\
\hline
Sandb. \etal \cite{Sandbach2012}& 6E, 60S, 6-CV, Win & 64.6\% \\
\hline
Fang \etal \cite{journalsivcFangZOSK12} & 6E, 100S, 10-CV, -- & 74.63\% \\
\hline
Le \etal \cite{Le2011FG} & 3E, 60S, 10-CV, Full seq. & 92.22\% \\
\hline
Xue \etal \cite{Xue2015} & 6E, 60S, 10-CV, Full seq. & 78.8\% \\
\hline
Berretti \etal \cite{Berretti2013} & 6E, 60S, 10-CV, Full seq. & 79.4\% \\
\hline
Berretti \etal \cite{Berretti2013} & 6E, 60S, 10-CV, Win=6 & 72.25\% \\
\hline
Ben Amor \etal \cite{benamor2014} & 6E, 60S, 10-CV, Full seq. & 93.21\% \\
\hline
Ben Amor \etal \cite{benamor2014} & 6E, 60S, 10-CV, Win=6. & 93.83\% \\
\hline
%\hline
%\textbf{This work SDM$_1$ }         & 6E, 60S, 10-CV, Apex. & \textbf{89.5}\% %\\
%\hline
%\textbf{This work SDM$_2$}          & 6E, 60S, 10-CV, Apex. & \textbf{90.62}\% %\\

\textbf{This work -- \textit{SVM} on $\bar{\chi}$}  & 6E, 60S, 10-CV, Full seq.& \textbf{93.39}\% \\
\hline
\textbf{This work -- \textit{HMM} on $\chi(t)$}  & 6E, 60S, 10-CV, Full seq. & \textbf{94.18}\% \\
%\hline
\end{tabularx}
\end{center}
\label{Table3}
\end{table}
%%%%%%%%%%%%%%%%%%%%%%%%%%%%%%%%%%%%%%%%%%%%%%%%%%%%%%%%%%%%%%%%%%%%%%%

Most of the results reported on BU-4DFE are shown in Table \ref{Table3}. In this table,
$\#$E means the number of expressions, $\#$S is the number of subjects, $\#$-CV provides the number of cross-validation used, \textit{Full Seq./Win} means the decision is made based on the analysis of the full sequence or on sub-sequences captured using a sliding window. The studies \cite{SunECCV2008, Sun2010} report one of the highest accuracy when using a sliding window of $6$ frames, nevertheless, the approach requires manual annotation of 83 landmarks on the first frame. Moreover, the vertex-level dense tracking scheme is time consuming. In a more recent work from the same group developed by Reale \etal \cite{fgrRealeZY13}, the authors propose 4D (\textit{Space-Time}) features termed \textit{Nebula} computed on a fixed-size window of $15$ frames. The best accuracy reported is 76.9\% using sequences of $100$ subjects, when the 3D video segmentation to limit the expressive time interval is performed manually. In \cite{journalsivcFangZOSK12}, Fang \etal obtained an accuracy of 74.63\% with $507$ sequences of $100$ subjects. Le \etal \cite{Le2011FG} evaluate their algorithm consisting of level curves and \textit{HMMs} on $60$ subjects sequences on three expressions (\textit{HA}, \textit{SA} and \textit{SU}) and display an accuracy of 92.22\%. It should be noted that the proposed approach is evaluated when considering full sequences which is a major difference to the works \cite{SunECCV2008,Sun2010,fgrRealeZY13,Sandbach2012}.
It is pointed out in \cite{Berretti2013} that the problem of the window-based evaluation protocol is to label all sub-sequences from the neutral intervals as one of the six expressions which can bias the final result.
Compared to the results listed in Table \ref{Table3}, the proposed approach outperforms existing approaches, where (1) no landmark detection is required; (2) no dimensionality reduction or feature selection techniques are applied; and (3) A vertex-level 4D dense registration and quantification of the deformations are led jointly through a Riemannian approach. The temporal filtering amplifies these deformations and consequently reveals hidden (subtle) 4D facial motions.
%%%%%%%%%%%%%%%%%%%%%%%%%%%%%%%%%%%%%%%%%%%%%%%%%%%%%%%%

%%%%%%%%%%%%%%%%%%%%%%%%%%%%%%%%%%%%%%%%%%%%%%%% Section 1. Conclusions.
\section{Conclusions}
\label{Concl}
In this paper, a spatio-temporal processing approach for effective 4D \textit{FER} is presented. 
It focus on an important issue -- expression magnification to reveal subtle deformations. After a preprocessing step, the flow of 3D faces is analyzed to capture the spatial deformations based on the Riemannian method where registration and comparison are achieved jointly. Then, the obtained deformations are amplified using the temporal filter over the 3D face video. The combination of these two ideas performs accurate vertex-level registration of 4D faces and reveal hidden (subtle) deformations in 3D facial sequences.
Experiments on BU-4DFE dataset demonstrate the effectiveness of the proposed method.

%\bibliographystyle{IEEEtran}
%\bibliography{IEEEabrv,TACBIB}



\end{document}